%% file: bare_jrnl_new_sample4.tex
\documentclass[lettersize,journal]{IEEEtran}
\usepackage{amsmath,amsfonts}
\usepackage{algorithmic}
\usepackage{algorithm}
\usepackage{array}
\usepackage[caption=false,font=normalsize,labelfont=sf,textfont=sf]{subfig}
\usepackage{textcomp}
\usepackage{stfloats}
\usepackage{url}
\usepackage{verbatim}
\usepackage{graphicx}
\usepackage{cite}
\usepackage{indentfirst}
\usepackage{bm}
\usepackage{multirow}
\usepackage{multicol}
\usepackage{ulem}
\usepackage{makecell}
\usepackage{booktabs}
\usepackage{bm}
\usepackage{bbm}
\usepackage{pifont}
\usepackage{subfloat}
\usepackage{xcolor}
\usepackage{hyperref}

\usepackage{algorithm}
\usepackage{algorithmic}

\newcommand{\unconfirm}[1]{\textcolor{black}{#1}}
\newcommand{\qiankun}[1]{\textcolor{black}{#1}}

\begin{document}

\title{MAFE R-CNN: Selecting More Samples to Learn Category-aware Features for Small Object Detection}
\author{Yichen Li, Qiankun Liu, Zhenchao Jin, Jiuzhe Wei, Jing Nie, Ying Fu,~\IEEEmembership{Senior Member,~IEEE}
\thanks{Yichen Li and Ying Fu are with School of Computer Science and Technology, Beijing Institute of Technology; (Email: {liyichen, fuying}@bit.edu.cn)}
\thanks{Qiankun Liu is with School of Computer and Communication Engineering, University of Science and Technology Beijing; (Email: liuqk3@ustb.edu.cn)}
\thanks{Zhenchao Jin is with School of Computer Science and Technology, University of Hong Kong; (Email: blwx96@connect.hku.hk)}
\thanks{Jiuzhe Wei is with Beijing Institute of Space Mechanics and Electricity; (Email: wjz\_casc@163.com)}
\thanks{Jing Nie is with School of information and electronics, Beijing Institute of Technology; (Email: 3420235028@bit.edu.cn)}
\thanks{This work was supported by the National Natural Science Foundation of China (62331006, 62171038, 62088101, and 62402042), the Fundamental Research Funds or the Central Universities}
}

\markboth{Journal of \LaTeX\ Class Files,~Vol.~14, No.~8, August~2021}%
{Shell \MakeLowercase{\textit{et al.}}: A Sample Article Using IEEEtran.cls for IEEE Journals}


\maketitle

\begin{abstract}
Small object detection in intricate environments has consistently represented a major challenge in the field of object detection.
In this paper, we identify that this difficulty stems from the detectors' inability to effectively learn discriminative features for objects of small size, compounded by the complexity of selecting high-quality small object samples during training, which motivates the proposal of the \uline{M}ulti-Clue \uline{A}ssignment and \uline{F}eature \uline{E}nhancement R-CNN.
Specifically, MAFE R-CNN integrates two pivotal components.
The first is the Multi-Clue Sample Selection (MCSS) strategy, in which the Intersection over Union (IoU) distance, predicted category confidence, and ground truth region sizes are leveraged as informative clues in the sample selection process. 
This methodology facilitates the selection of diverse positive samples and ensures a balanced distribution of object sizes during training, thereby promoting effective model learning.
The second is the Category-aware Feature Enhancement Mechanism (CFEM), where we propose a simple yet effective category-aware memory module to explore the relationships among object features. Subsequently, we enhance the object feature representation by facilitating the interaction between category-aware features and candidate box features.
Comprehensive experiments conducted on the large-scale small object dataset SODA validate the effectiveness of the proposed method. The code will be made publicly available.
\end{abstract}


\section{Introduction}
\IEEEPARstart{S}mall Object Detection (SOD) aims to classify and locate the objects with limited pixels, which plays a crucial role in various scenarios, such as autonomous driving, remote sensing surveillance, drone scene analysis, and maritime rescue~\cite{chen2023attentive,cheng2023towards,yu2020scale,xu2022rfla,cao2021visdrone,chen2024accurate,liu2024infrared,li2023sliding}.
Specifically, in these scenarios, the key information is contained in small objects.
The loss or poor detection of small objects may lead to management issues or even dangerous incidents~\cite{balasubramaniam2022object}.
However, identifying small objects in images is difficult. Detectors are limited by the extremely small size of these objects, making it difficult to achieve satisfactory detection performance.

\begin{figure}[t]
    \scriptsize
    \centering
    \begin{minipage}[b]{0.32\linewidth}
        \centering
        \includegraphics[width=\linewidth]{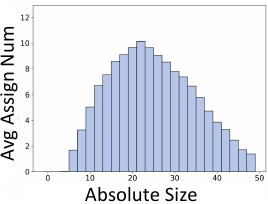}
        \text{Euclidean-Distance\cite{xu2021dot}} \\ 
    \end{minipage}
    \begin{minipage}[b]{0.32\linewidth}
        \centering
        \includegraphics[width=\linewidth]{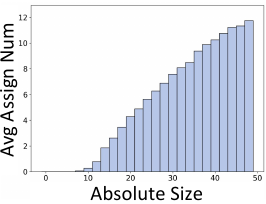}
        \text{IoU-Distance\cite{ren2015faster}} \\ 
    \end{minipage}
    \begin{minipage}[b]{0.32\linewidth}
        \centering
        \includegraphics[width=\linewidth]{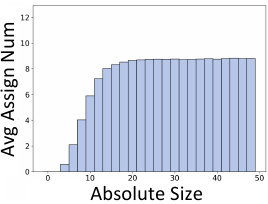}
        \text{MCSS(Ours)} \\ 
    \end{minipage}
    \vspace{1em} 
    \small
    \textbf{(a) Statistical results of different sample selection methods}
    \scriptsize
    \begin{minipage}[t]{0.23\linewidth}
        \centering
        \includegraphics[width=\linewidth]{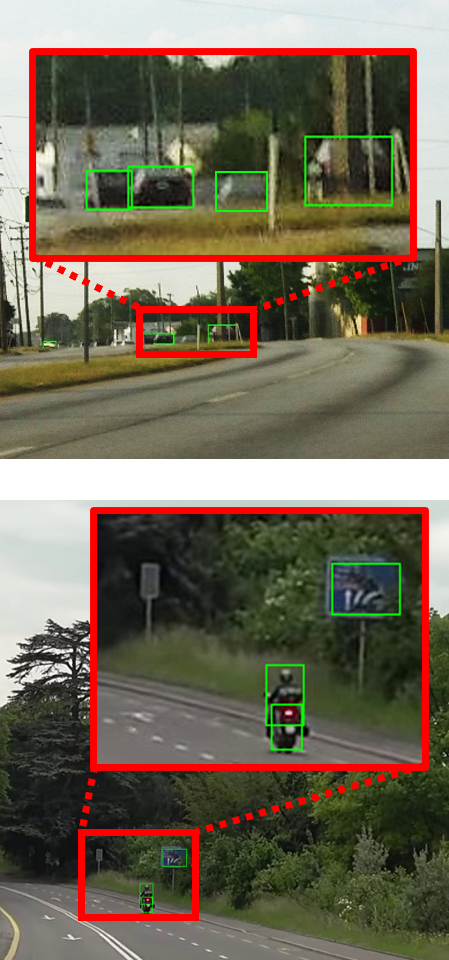}
        \text{Ground Truth} \\ 
    \end{minipage}
    \begin{minipage}[t]{0.23\linewidth}
        \centering
        \includegraphics[width=\linewidth]{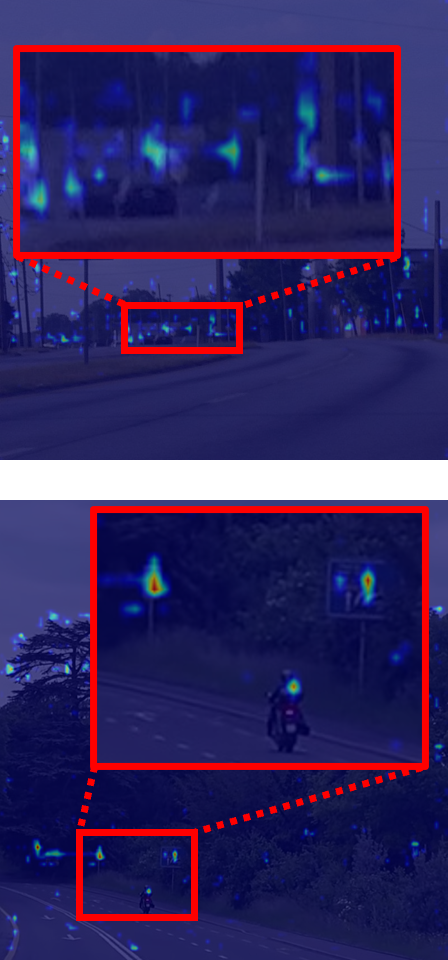}
        \text{Cascade R-CNN\cite{cai2019cascade}} \\ 
    \end{minipage}
    \begin{minipage}[t]{0.23\linewidth}
        \centering
        \includegraphics[width=\linewidth]{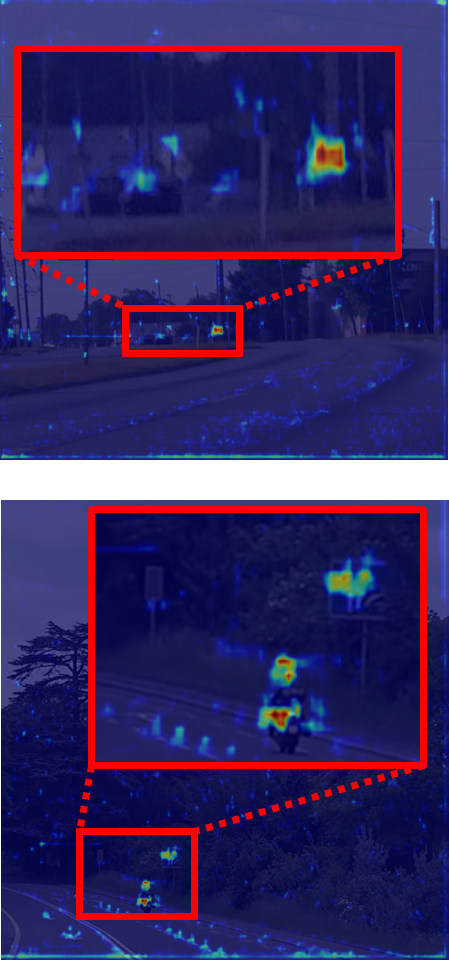}
        \text{CFINet\cite{yuan2023small}} \\ 
    \end{minipage}
    \begin{minipage}[t]{0.23\linewidth}
        \centering
        \includegraphics[width=\linewidth]{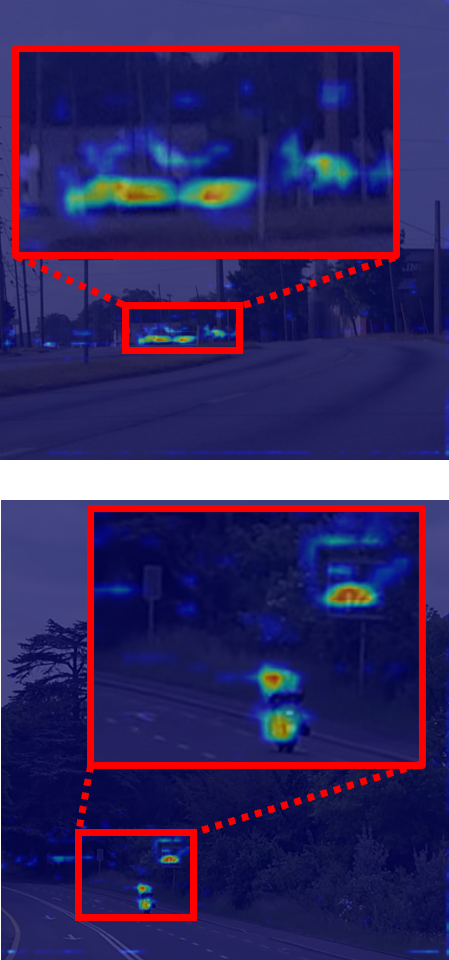}
        \text{MAFE R-CNN(Ours)} \\ 
        \vspace{1em}
    \end{minipage}
    \vspace{1em} 
    \small
    \textbf{(b) Feature visualization of different methods}
    \vspace{-0.3cm}
    \caption{(a) The average number of samples assigned by different assigners for objects of various sizes, where absolute size corresponds to the square root of the object area. (b) Feature map visualization. Brighter colors indicate the higher attention of the model to the region. Traditional methods exhibit low attention to small objects, making it difficult for the model to acquire small object samples and high-quality small object features. The proposed MAFE R-CNN shows a more balanced sample allocation and higher-quality small object features, contributing to more effective small object prediction.}
    \label{intro_fig}
\end{figure}

The difficulty in accurately detecting small objects mainly arises from imbalanced samples and blurred features:
(1) \textbf{Imbalanced samples}. As shown in Figure~\ref{intro_fig}(a), fixed strategies and thresholds lead to imbalanced sample assignment. Tiny prediction shifts significantly affect the sample distribution of small objects. This prevents small objects from being well-optimized during training, leading to the detector's inability to predict them.
(2) \textbf{Blurred features.} As shown in Figure~\ref{intro_fig}(b), small objects typically contain only a few pixels, making their features difficult to be discriminative. Feature maps containing more noise can obscure small object features, adversely affecting the detector's prediction of small objects.

To address the first issue, some works~\cite{zhang2017s3fd,zhu2018seeing,xu2021dot,xu2022rfla} propose sample assignment methods. 
These methods aim to provide the model with sufficient positive samples for small objects by improving the assignment strategy. 
They only rely on predicted position information to select samples, which can easily introduce low-quality samples and often fail to match samples when dealing with extremely small objects, leading to inadequate samples. 
Specifically, as the object area decreases, even tiny shifts can significantly affect the similarity between sample and the ground truth.
Moreover, predefined thresholds are inflexible for various object sizes. 
Therefore, positional information alone cannot serve as the sole criterion for determining positive and negative samples, and optimized sample selection methods need to be designed.

To address the second issue, some methods introduce scale-specific detectors, where features at different layers are responsible for detecting objects of corresponding sizes~\cite{li2017scale,lin2017feature,yang2022querydet}. Other methods aim to obtain higher-quality small object features through techniques such as image super-resolution~\cite{bai2018finding,bai2018sod,noh2019better,deng2021extended}, feature fusion~\cite{liu2018path,zhao2019m2det,gong2021effective}, and feature imitation~\cite{wu2020self,kim2021robust,yuan2023small,liu2024mlfa}. 
These methods usually require complex network structures (\textit{e.g.}, super-resolution networks~\cite{deng2021extended}, feature imitation loss~\cite{kim2021robust, yuan2023small}), and tend to focus more on feature information within the image or the relationship between small and large objects to enhance the discriminability of small object features. 
They often overlook the similarity between features of similar objects and the contextual semantic connections among objects.
The essence of deep learning-based object detection is to classify and regress similar feature regions across all images, rather than performing independent classification and regression within each individual image.
Therefore, to achieve accurate prediction in small objects, it is important to consider the feature information shared among similar objects.

In this paper, we propose the \underline{\bf M}ulti-clue \underline{\bf A}ssignment and \underline{\bf F}eature \underline{\bf E}nhancement (MAFE R-CNN) to address the issues of imbalanced samples and blurred features for small object detection. There are two pivotal components, \textit{i.e.}, Multi-Clue Sample Selection (MCSS) and Category-aware Feature Enhancement Mechanism (CFEM). 
(1) {\bf MCSS}: 
We design a multi-clue matching strategy to optimize sample assignment. 
\unconfirm{
During training, the positive samples are selected with a dynamic assignment threshold.
The IoU-distance, category confidence, and ground truth sizes are considered, which facilitates the selection of more diverse positive samples and ensures a balanced distribution across different object sizes, promoting the detector to learn to detect small objects effectively.
}
(2) {\bf CFEM}: 
We introduce a simple yet effective category-aware memory to explore the relationships among object features. The memory module is dynamically updated based on the ground truth region features during training and is fixed during inference. 
To enhance the candidate box features through the interaction with the memory features, we first perform pre-classification on each candidate box to obtain confidence scores as the weight to compute a weighted sum of memory features, resulting in the category-aware feature of the candidate box. 
Then, these category-aware features are interacted with the original candidate box features through cross-attention, enhancing object features for classification and regression.
Finally, these category-aware features interact with the original candidate box features through cross-attention, enhancing object features for classification and regression.

The main contributions of this work are summarized as:
\begin{itemize}
    \item \unconfirm{We propose Multi-clue Assignment and Feature Enhancement R-CNN (MAFE R-CNN) that incorporates multi-clues and category information into sample selection and feature enhancement, respectively, offering a more effective solution for the SOD task.}
    \item We design a Multi-Clue Sample Selection (MCSS) strategy. Through multi-clue matching and dynamic selection, MCSS provides more balanced and high-quality positive samples for small objects during the training phase.
    \item We introduce a Category-aware Feature Enhancement Mechanism (CFEM), which addresses the challenge of feature extraction by enhancing them through a category-aware memory module and information interaction.
\end{itemize}
\unconfirm{Experimental results show that MAFE R-CNN achieves state-of-the-art detection performance on the large-scale small object dataset SODA.}
\section{Related Work}

This section briefly reviews related works for small object detection from different aspects, including small object detection, feature memory module, and sample selection.

\subsection{Small Object Detection}

Over the past decade, object detection has become a popular research field, reaching a high level of performance. Various types of detectors~\cite{girshick2014rich,duan2019centernet,zhu2020deformable,ge2021yolox,jin2020safnet,liu2024siamese,chen2024frequency} have emerged, typically extracting object features from a backbone network and making decisions through a detection head. 
However, this detection paradigm struggles with SOD tasks due to the limited size of small objects, leading to significant loss of feature information and making detection difficult. 
To address the problem, existing small object detection methods introduce some strategies to achieve better performance in SOD tasks.

Some studies in~\cite{li2017scale,lin2017feature,yang2022querydet} use Feature Pyramid Network (FPN) and parallel branches for multi-scale prediction, leveraging high-resolution features for small object detection.
Other methods such as PANet~\cite{liu2018path}, StairNet~\cite{woo2018stairnet}, IPG-Net~\cite{liu2020ipg} and CFP~\cite{quan2023centralized} employ feature fusion techniques to facilitate information interaction across different feature layers, achieving better feature representation for small objects. 
Additionally, more researchers attempt to enhance small object features by imitating the features of larger objects~\cite{wu2020self}.
On one hand, Perceptual GAN~\cite{li2017perceptual} designs a generator to optimize high-quality representations of small objects, aiming to deceive the discriminator. 
The study~\cite{bai2018finding} proposes a novel process to restore blurry small objects into clearer targets for recognition.
Noh \textit{et al.}~\cite{noh2019better} enhance small objects via image super-resolution, improving the feature representation quality of small objects. 
On the other hand, Wu \textit{et al.}~\cite{wu2020self} and Kim \textit{et al.}~\cite{kim2021robust} apply similarity learning to bring small object features closer to those of larger objects. CFINet~\cite{yuan2023small} introduces a feature imitation method that completes small object features through imitation and incorporates a feature imitation loss.

Despite effective progress in small object detection, most methods are still limited by features within a single image and the relationship between small and large objects. This work exploits the connections between features of the same category and explores potential category-aware features between objects to enhance detection performance.

\subsection{Memory Module}

{The memory module is a key component that enables the model to acquire and retain effective information beyond the input image, facilitating information aggregation across multiple images during training. This capability is particularly evident in various computer vision tasks~\cite{kumar1994mega,jin2022mcibi++,gramfort2014mne,wang2020cross,chen2020memory,zhong2019invariance,li2019memory,wu2019long,zhang2023object,liu2024transformer}.} 
A notable example is MEGA~\cite{chen2020memory}, which employs a Long-Range Memory (LRM) module to store global semantic features, thereby enhancing image features in keyframes for video object detection. 
Similarly, MCIBI$++$~\cite{jin2022mcibi++} improves pixel-level feature representation in segmentation tasks by learning an image memory module. 
In the field of few-shot learning, MNE~\cite{gramfort2014mne} proposes using memory-based neighborhood embedding to enhance general convolutional features. 
Additionally, in deep metric learning for image retrieval, XBM~\cite{wang2020cross} introduces a cross-batch memory module specifically designed to mine informative examples across multiple small batches, thereby improving overall model performance.

Previous methods~\cite{wang2020cross,chen2020memory,jin2022mcibi++} construct memory units for image-level features, which lack focus on specific objects. In contrast, we introduce an instance-level feature memory module to learn contextual information among objects, thereby enhancing the representation ability of low-quality object features during the inference process.

\begin{figure*}[t]
	\centering
	\includegraphics[width=\linewidth]{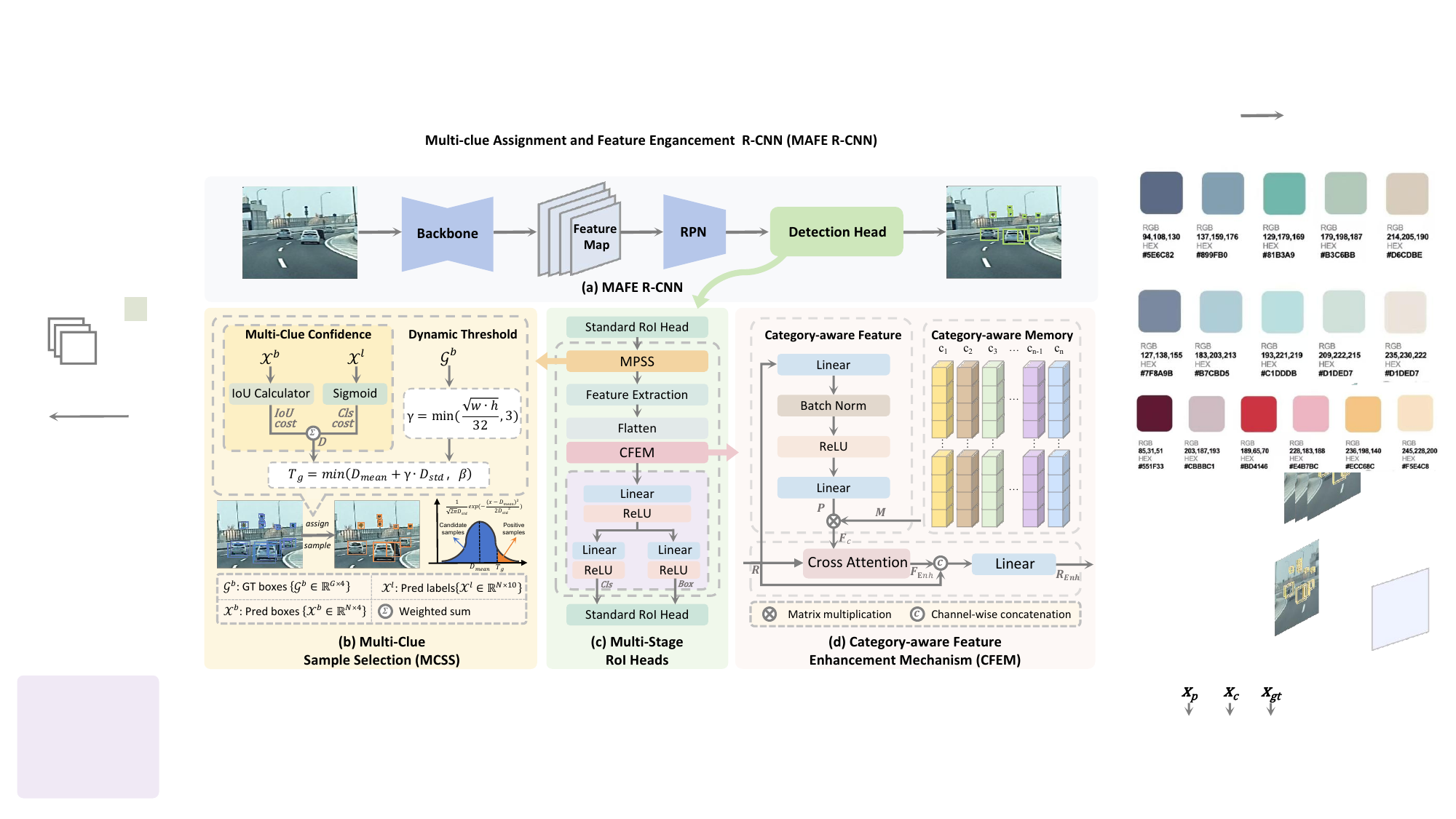}
	\caption{Illustration of \qiankun{the proposed MAFE R-CNN}. 
    Our method integrates Multi-Clue Sample Selection (MCSS) and Category-aware Feature Enhancement Mechanism (CFEM) into the MAFE R-CNN for training. During inference, MCSS is not used for sample selection, and the category-aware memory remains fixed. The three-stage prediction process based on the input features of small objects produces the final prediction results.}
	\label{model_structure}
\end{figure*}

\subsection{Sample Selection}

During object detection training, model loss is constrained by both positive and negative samples. When small objects are involved, the division of positive and negative samples becomes more challenging. Predictions with a high overlap with ground truth are assigned as positive samples for object regression loss calculation, while negative samples refer to completely incorrect predictions, which are only used in classification loss calculation.  
Traditional detectors~\cite{girshick2014rich,ren2015faster,cai2019cascade} typically use overlap-based or distance-based matching strategies to select samples and rely on manually designed thresholds as decision criteria.  
For small objects, $S^3$FD~\cite{zhang2017s3fd} increases the number of smaller anchor boxes to improve recall rates. Zhu \textit{et al.}~\cite{zhu2018seeing} account for anchor strides in calculating overlaps and propose the Expected Maximum Overlap (EMO) score. 
Similarly, RFLA~\cite{xu2022rfla} measures the similarity between the Gaussian receptive field of each feature point and the ground truth in label assignment. ATSS~\cite{zhang2020bridging} automatically selects positive and negative training samples based on the statistical characteristics of the objects. 

\unconfirm{In this work, we aim to balance the number of positive samples across different object sizes while selecting high-quality positive samples. To achieve this, we combine \qiankun{the clues of} IoU-distance, category information, and object size for a more comprehensive sample selection.}
\section{\qiankun{MAFE R-CNN}}
In this section, we first introduce the motivation and overview of our method, then provide a detailed description of core designs, \textit{i.e.}, multi-clue sample selection (MCSS), and category-aware feature enhancement mechanism (CFEM). Finally, we illustrate how to integrate them into the MAFE R-CNN and the training details of the proposed method.

\subsection{Motivation and Network Architecture}

Despite the promising performance in small object detection methods~\cite{xu2022rfla,yuan2023small}, challenges in detection performance persist due to imbalanced sample assignment and blurred features. 
Balanced samples better optimize model parameters, while high-quality features facilitate object classification and regression. 
Therefore, designing small object detection methods requires optimized sample selection and feature extraction.


\subsubsection{Inadequate Positive Sample During Training}
During the training process of the detector, we need to classify predictions into positive and negative samples at each stage, and then perform subsequent loss calculations. However, the final regression loss calculation only involves positive samples, so this step is crucial for training the detector. 
The scarcity or imbalance of training samples can prevent the detector from learning high-quality small object features, potentially causing the model to lose the ability to detect objects in low-sample data.
Traditional sample selection methods use Euclidean distance~\cite{xu2021dot} or IoU distance~\cite{girshick2015fast} as criteria for positive and negative sample selection. 
However, these are highly sensitive to sample size. 
As shown in Figure~\ref{intro_fig}(a), Euclidean distance only captures samples within a specific size range, while IoU distance restricts the model's ability to acquire small-object samples. 
To avoid this issue, rather than focusing on tuning parameters, our aim is to use a more flexible algorithm to control sample selection, thereby capturing more small-object samples and achieving a more balanced sample assignment.

\subsubsection{Model Lacks Attention to Small Object Features}
Another challenge of traditional object detection methods~\cite{ren2015faster,cai2019cascade} is the lack of attention to small-object features, especially in scenarios with a high proportion of small objects. 
First, since the model does not pay enough attention to small objects, feature fusion methods~\cite{liu2018path,woo2018stairnet,liu2020ipg} do not fundamentally resolve the issue of small-object features being overshadowed, and they instead increase computational costs. 
Additionally, shared feature spaces constructed through similarity learning~\cite{wu2020self,kim2021robust,yuan2023small} are not optimal, as shown in Figure~\ref{intro_fig}(b), where more feature noise is introduced. 
In contrast, this paper posits that features among objects of the same category are more closely related. 
In order to obtain more accurate feature information, it is crucial to consider the features of the same category objects in other images.

\subsubsection{Overview of Our Method}
{To address the challenges of small object detection, we propose Multi-clue Assignment and Feature Enhancement (MAFE R-CNN). 
The architecture of the proposed method is shown in Figure~\ref{model_structure}(a). 
Similar to other two-stage detection frameworks, the network consists of a backbone, a Region Proposal Network (RPN), and a detection head with multi-stage RoI heads for classification and regression. 
To obtain more high-quality positive samples for small objects, we design a Multi-Clue Sample Selection (MCSS) module (Section~\ref{MCSS}), which maintains balance in sample selection by considering multiple factors. 
Next, we propose a Category-aware Feature Enhancement Mechanism (CFEM) (Section~\ref{CFEM}), which explores the potential category information among similar object features. 
Finally, we integrate these two modules into the Category-aware Detection Head (see Section~\ref{MAFE}) and further explain their interactions and training strategies in detail.}

It is worth noting that the proposed method fundamentally differs from conventional multi-stage RoI head detectors~\cite{cai2019cascade}. 
In previous methods, the IoU threshold at each stage is manually set based on empirical rules, resulting in inflexibility that tends to filter out more small objects, 
and makes it difficult to focus on small object features. 
This leads to a severe imbalance between large and small objects during training, ultimately diminishing the detector’s ability to recognize small objects. 
In contrast, the proposed method aims to flexibly select samples across multiple stages and achieve more precise object detection through feature enhancement.

\subsection{\qiankun{Multi-Clue Sample Selection}}
\label{MCSS}

In this part, we introduce MCSS as shown in Figure~\ref{model_structure}(b). 
The essence of MCSS is to select higher-quality samples for classification and regression. 
It consists of two main steps: generation of multi-\qiankun{clue} confidence and dynamic threshold calculation based on region size and distribution. The positive and negative samples are selected based on the dynamic threshold. 
We first present the design of MCSS. Then, we discuss in detail the process of generating multi-\qiankun{clue} confidence. Finally, we describe the dynamic threshold based on region size and distribution.
\subsubsection{Design of MCSS}
Algorithm~\ref{MCSS_algorithm} describes how the proposed method selects positive and negative samples. 
Each input consists of $G$ ground truth bounding boxes and labels from a single image, denoted as $\{ \mathcal{G}^{b} \in \mathbb{R}^{G \times 4}, \mathcal{G}^{l} \in \mathbb{R}^G \}$, along with $N$ predicted bounding boxes and label scores, denoted as $\{ \mathcal{X}^{b} \in \mathbb{R}^{N \times 4}, \mathcal{X}^{l} \in \mathbb{R}^{N \times C} \}$,  where C is the numbner of categories.
For each ground truth $\mathcal{G}_g$ in the current image, we identify positive samples among the candidate samples.
First, we select the $k$ candidate samples $\hat{\mathcal{X}}^b \in \mathbb{R}^{k \times 4}$ that are closest to the center of the ground truth $\mathcal{G}_g$ based on the Euclidean distance. 
Next, we compute the category confidence $\mathcal{D}_c \in \mathbb{R}^k$ and the IoU distance $\mathcal{D}_{iou} \in \mathbb{R}^k$ for each candidate sample relative to $\mathcal{G}_g$. 
Then, based on the weight $\alpha$, we calculate the multi-clue confidence $D \in \mathbb{R}^k$ for these candidate samples with respect to $\mathcal{G}_g$.
Afterward, we calculate the sample threshold for $\hat{\mathcal{X}}^b$ relative to $\mathcal{G}_g$ based on the sample distribution\unconfirm{, \textit{i.e.}, the mean $m_{\mathcal{D}}$ and standard deviation $v_{\mathcal{D}}$ of $\mathcal{D}$.}
Subsequently, we calculate the area of $\mathcal{G}_g$, denoted as $S_g$, and use it to compute the standard ratio $\gamma$. 

Finally, we select the candidate samples with multi-clue confidence greater than or equal to the threshold $T_g$ as the final positive samples. Additionally, following standard practices~\cite{girshick2015fast}, if a candidate sample is assigned to multiple ground truth labels, we choose the multi-clue sample with the highest confidence as the positive sample, while the others are treated as negative samples.

\begin{algorithm}[t]
\caption{Multi-Clue Sample Selection}
\label{MCSS_algorithm}
\begin{algorithmic}[1]
    \REQUIRE
        \par ground truth boxes $\mathcal{G}^{b}$, labels $\mathcal{G}^{l}$ in the image 
        \par \hspace{1.2em} predicted boxes $\mathcal{X}^{b}$, label scores $\mathcal{X}^{l}$ in the image 
        \par \hspace{1.2em} hyperparameter related to sample selection $k$
        \par \hspace{1.2em} category confidence weight $\alpha$
        \par \hspace{1.2em} minimum positive sample threshold $\beta$
    \ENSURE
        \par positive Samples $\mathcal{P}$
        \par \hspace{1.2em} negative samples $\mathcal{N}$
    \FOR{each ground truth $\mathcal{G}_g$ in $\mathcal{G}$}
        \STATE select $k$ candidate boxes from $\mathcal{X}^{b}$ whose center are closest to the center of $\mathcal{G}_g$: $\hat{\mathcal{X}}$
        \STATE compute category confidence $\mathcal{D}_c$ between $\hat{\mathcal{X}}^{l}$ and $\mathcal{G}_g^{l}$
        \STATE compute IoU $\mathcal{D}_{iou}$ between $\hat{\mathcal{X}}^{b}$ and $\mathcal{G}_g^{b}$
        \STATE compute multi-clue confidence matrix: $\mathcal{D}$ = $\alpha \cdot \mathcal{D}_c + (1-\alpha) \cdot \mathcal{D}_{iou}$
        \STATE compute mean of $\mathcal{D}$: $m_\mathcal{D}$ = $Mean(\mathcal{D})$
        \STATE compute standard deviation of $\mathcal{D}$: $v_\mathcal{D}$ = $Std(\mathcal{D})$
        \STATE compute ground truth absolute area $S_{g}$ = $\sqrt{w \cdot h}$
        \STATE compute standard ratio $\gamma$ = $min\{S_{g}/S_{max}, 3\}$
        \STATE compute threshold for $\mathcal{G}_g$: $T_g$ = $min( m_\mathcal{D} + \gamma \cdot v_\mathcal{D}, \beta)$
        \FOR{each sample $s$ in $\hat{\mathcal{X}}$}
            \IF{$\mathcal{D}_s \geq T_g$ and center of $s$ in $\mathcal{G}^b_g$}
                \STATE $\mathcal{P} \cup s$ 
            \ENDIF
        \ENDFOR
    \ENDFOR
    \STATE $\mathcal{N}$ = $\mathcal{X} - \mathcal{P}$
    \RETURN $\mathcal{P}, \mathcal{N}$
\end{algorithmic}
\end{algorithm}

\subsubsection{Multi-\qiankun{Clue} Confidence} 
Considering positional and feature information is an effective approach to obtaining high-quality samples. 
The IoU distance $\mathcal{D}_{iou}$ between samples and the ground truth provides positional information. 
A higher $\mathcal{D}_{iou}$ indicates that the candidate sample is closer to the corresponding ground truth box. 
However, due to the tiny size of objects, small objects typically have a lower $\mathcal{D}_{iou}$. 
This does not necessarily mean that the quality of small object samples is low. 
Using IoU-distance to evaluate sample quality essentially aims to ensure that the candidate sample overlaps as much as possible with the ground truth, which implies a higher degree of feature similarity between them. This allows the trained detector to make better predictions. 
Thus, we aim to represent the feature quality of these samples by calculating the category confidence $\mathcal{D}_c$:
\begin{equation}
    \mathcal{D}_c = \text{Sigmoid}(\hat{\mathcal{X}}^{l}[:, \mathcal{G}_g^{l}]), 
\end{equation}
which indicates the confidence that the samples category is the same as the ground truth category.
A higher $\mathcal{D}_c$ indicates a higher similarity between the sample and the ground truth box, meaning that $\mathcal{D}_c$ provides a more intuitive representation of feature quality. 
Therefore, considering the nature of object detection, we combine the IoU distance and category confidence to compute the multi-\qiankun{clue} confidence as a factor for evaluating the quality of candidate samples:
\begin{equation}
    \mathcal{D} = \alpha \times \mathcal{D}_c + (1-\alpha) \times \mathcal{D}_{iou},
\end{equation}
where $\alpha$ is the weight of the category confidence. 
High-quality feature information aids classification, while better location information helps regression. 
By combining both position and feature information, the resulting multi-\qiankun{clue} confidence can more objectively assess the sample quality.

\subsubsection{Dynamic threshold} 
MCSS calculates dynamic threshold based on the ground truth region size and sample distribution. 
The confidence of candidate samples approximates a normal distribution, and typically around $0.2k$ candidate samples are assigned as positive samples~\cite{zhang2020bridging}.
The multi-clue confidence mean $m_\mathcal{D}$ represents the quality of the candidate samples, while the standard deviation $v_\mathcal{D}$ reflects the stability of this quality. 
Based on the mathematical properties of the normal distribution, using $m_\mathcal{D} + v_\mathcal{D}$ as the threshold results in approximately $0.16k$ positive samples.
However, due to the object's region size, larger objects have more candidate samples. Maintaining the same proportion does not yield a balanced number of samples. 
To ensure fairness in sample quantity, we design a region-based standard ratio calculation method. This method computes the standard ratio $\gamma$ based on the region size of the ground truth:
\begin{equation}
    \gamma = \min\left\{ \frac{S_{g}}{S_{max}}, 3 \right\},
\end{equation}
where $S_{max}$ represents the maximum absolute size of small objects, 
and the maximum value for $\gamma$ is set to $3$.
We then calculate the dynamic threshold \(T_g\) as:
\begin{equation}
    T_g = \min( m_D + \gamma \cdot v_D, \beta).
    \label{formula_tg}
\end{equation}
By adjusting the threshold based on the mathematical distribution and region size of each ground truth, this method ensures a balanced number of positive samples across different object sizes.
It is worth noting that we set a minimum positive sample threshold $\beta$ to constrain the $T_g$. This is to avoid including low-quality samples generated by some challenging objects.

\subsection{Category-aware Feature Enhancement Mechanism}
\label{CFEM}

\begin{figure}[!t]
\centering
\includegraphics[width=\linewidth]{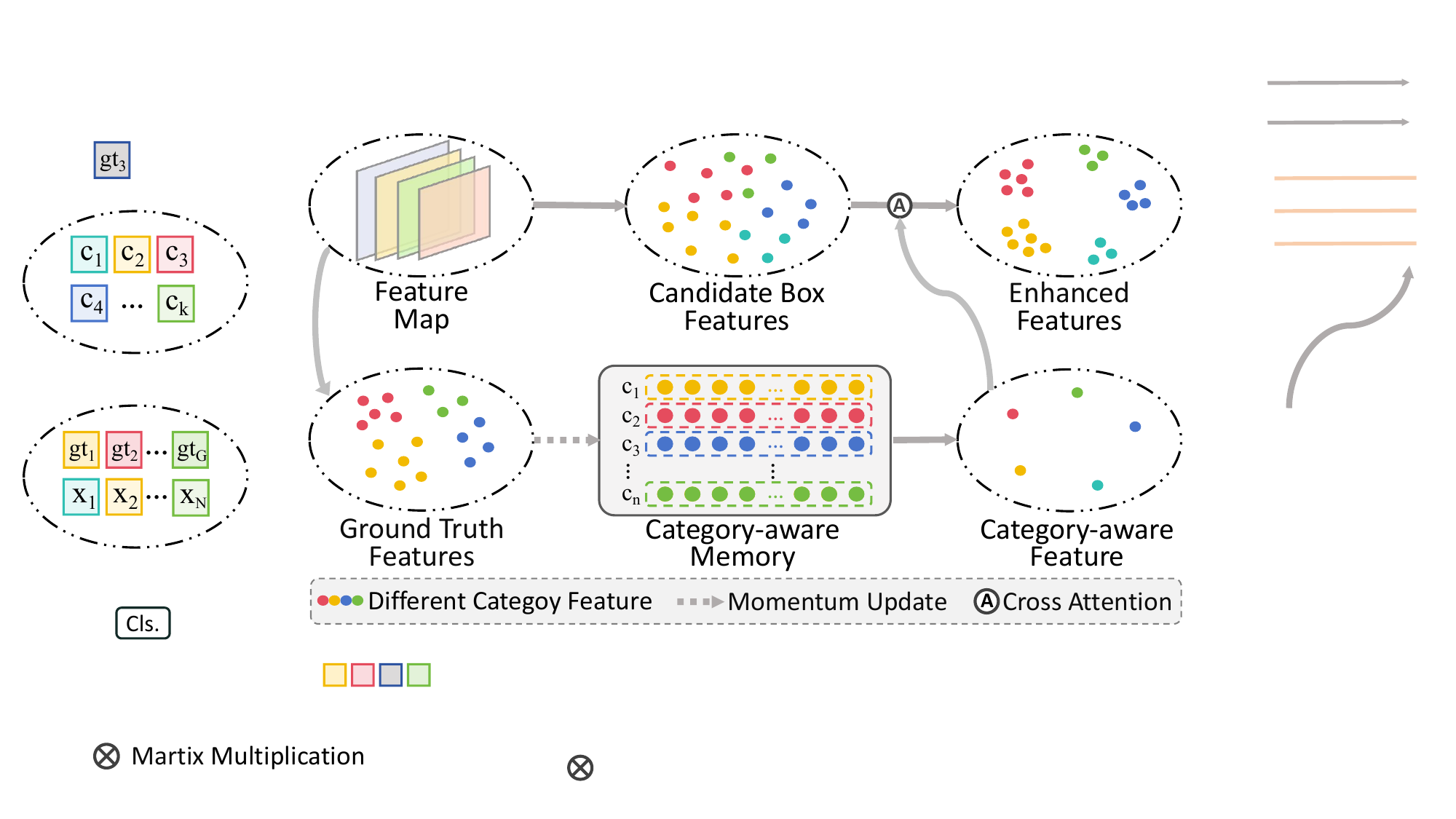}
\caption{The illustration of CFEM training pipeline. Ground truth features update the category-aware memory, which is used for feature enhancement.}
\label{cfem_fig}
\end{figure}

CFEM comprises three core components: Category-aware Memory, Category-aware Feature Generation, and Feature Interaction Enhancement.
As shown in Figure~\ref{cfem_fig}, the category-aware memory is updated with ground truth features after each iteration, resulting in an updated category-aware memory for feature enhancement in the next iteration.
The inference process is shown in Figure~\ref{model_structure}(d), we firstly employ a linear classifier to predict the category probabilities $\textit{\textbf{P}}  \in \mathbb{R}^{N \times C}$ for candidate boxes, where $C$ is the number of categories.
Then, we utilize $\textit{\textbf{P}}$ and the category-aware memory $\textit{\textbf{M}} \in \mathbb{R}^{C \times D}$ to generate category-aware features $\textit{\textbf{F}}_c \in \mathbb{R}^{N \times D}$ for the current candidate boxes, where $D$ represents the dimension of the category-aware feature.
Subsequently, we employ a cross-attention to enable information interaction between the candidate box features $\textit{\textbf{R}} \in \mathbb{R}^{N \times D}$ and $\textit{\textbf{F}}_c$, thereby supplementing the missing feature information in $\textit{\textbf{R}}$ and obtaining the enhanced feature $\textit{\textbf{F}}_{Enh}$.
Finally, we combine $\textit{\textbf{F}}_{Enh} \in \mathbb{R}^{N \times D}$ with $\textit{\textbf{R}}$ as the feature representation $\textbf{R}_{Enh} \in \mathbb{R}^{N \times D}$ of the candidate box for prediction.

In the following, we begin by introducing the learning process of category-aware memory during training. Then, we provide a detailed explanation of the category-aware feature generation. Finally, we explain how category-aware features interact with original features to enhance object representation.

\subsubsection{Learning of Category-aware Memory}
After each training iteration, we update the category-aware memory feature by conducting a weighted moving average operation:
\begin{equation}
\label{update_memory}
   \textit{\textbf{M}}^{t} = (1 - m)  \textit{\textbf{M}}^{t-1} + m \hspace{1pt} \mathcal{T} \hspace{-2pt} (\textit{\textbf{G}})^{t-1},
\end{equation}
where $t$ represents the current iteration number, $m$ denotes the momentum, and $\mathcal{T}(\textit{\textbf{G}}^{t-1}) \in \mathbb{R}^{C\times D}$ is the feature of all categories to update the features in the memory. Take the category $C_j$ for example, to ensure the generalization performance of the category-aware feature memory, we assume category $C_{j}$ corresponds to $K$ ground truth boxes features and update the memory feature for $C_{j}$. Then we compute the cosine similarity $\textit{\textbf{W}}_{cos_j} \in \mathbb{R}^{K \times 1}$ between the ground truth boxes feature $\textit{\textbf{G}} \in \mathbb{R}^{K \times D}$ and the memory feature $\textit{\textbf{M}}_{C_{j}} \in \mathbb{R}^{1 \times D}$:
\begin{equation}
        \textit{\textbf{W}}_{{cos_j}} = \frac{\textit{\textbf{G}} \cdot \textit{\textbf{M}}_{C_{j}}^{T}}{\|\mathbf{\textit{\textbf{G}}}\|_2 \cdot \| \textit{\textbf{M}}_{C_{j}}\|_2},
\end{equation}
where $\textit{\textbf{M}}_{C_{j}}$ represents memory feature belong to category ${C_{j}}$ and $\|\mathbf{\cdot}\|_2$ represents the L2 norm.
Finally, the feature used to update the memory feature belonging to the category $C_{j}$ (denoted as $\mathcal{T} \hspace{-2pt} (\textit{\textbf{G}})_j \in \mathbb{R}^{1\times D}$) is obtained through a weighted average based on $\textit{\textbf{W}}_{cos_j}$:
\begin{equation}
    \mathcal{T} \hspace{-2pt} (\textit{\textbf{G}})_j = \sum_{k=1}^K \frac{1-\textit{\textbf{W}}_{cos_{j,k}}}{{\sum_{k^{'}=1}^K (1-\textit{\textbf{W}}_{cos_{j,k^{'}}}})} \textit{\textbf{G}}_{k},
\end{equation}
where 
$\textit{\textbf{W}}_{{cos}_{j,k}}$ represents the cosine similarity between $\textit{\textbf{M}}_{C_{j}}$ to the $k^{th}$ ground truth feature in $\textit{\textbf{G}}$. It is worth noting that as the cosine similarity between the input features and the memory features increases, the proportion of their contribution to updating the memory features decreases. 
This weight assignment effectively prevents the memory features from overfitting to partially similar objects, thus enhancing the robustness of the feature memory. 
Regarding the background class feature distribution in the category-aware memory, we update it with the features of the two negative samples with the least overlap with all ground truth in the image.

\subsubsection{Generation of Category-aware Feature}
The detector performs classification and regression on the object based on the extracted features. 
Previous methods have enhanced small object features using feature fusion within the image or feature priors from larger objects~\cite{li2017scale,lin2017feature,yang2022querydet,deng2021extended,bai2018finding}.
In contrast, we delve deeper into the characteristics of small objects by leveraging the potential correlations between their features. 
The given input $\hat{\textit{\textbf{R}}} \in \mathbb{R}^{N \times 256 \times 7 \times 7}$ is obtained by extracting features from the previous stage's prediction results. We firstly flatten the features $\hat{\textit{\textbf{R}}}$ along the channel and spatial dimensions, followed by a linear layer and a ReLU activation to obtain the $\textit{\textbf{R}}$:
\begin{equation}
    \textit{\textbf{R}} = \text{ReLU}(\text{Linear}(\text{Flatten}(\hat{\textit{\textbf{R}}}))).
\end{equation}
To generate category-aware features for each candidate box based on its features, we utilize a linear classifier to predict the category of $\textit{\textbf{R}}$, obtaining the category probabilities $\textit{\textbf{P}}$ for each candidate box. Then, we use $\textit{\textbf{P}}$ as the weights for the memory features to perform weighted averaging, yielding the category-aware features $\textit{\textbf{F}}_c$ of the candidate boxes
\begin{equation}
    \textit{\textbf{F}}_c = \textit{\textbf{P}} \cdot \textit{\textbf{M}}.
\end{equation}

\subsubsection{Feature  Interaction Enhancement} 
After obtaining the category-aware features of the candidate boxes, the next step is to enhance the original candidate box features with $\textit{\textbf{F}}_c$. The cross-attention mechanism~\cite{chen2021crossvit} serves as an effective structure of feature interaction. Here, we treat $\textit{\textbf{R}}$ as the query and $\textit{\textbf{F}}_c$ as the key and value. 
Through cross-attention, we supplement the missing feature information in the original candidate boxes features, resulting in the enhanced features $\textit{\textbf{F}}_{Enh}$.
Finally, we concatenate $\textit{\textbf{F}}_{Enh}$ with $\textit{\textbf{R}}$ along the last dimension and pass them through a linear layer to form the new candidate box features representation $\textit{\textbf{R}}_{Enh}$, which are then fed into the box head for classification and regression:
\begin{equation}
    \textit{\textbf{R}}_{Enh} = \text{Linear}(\text{Concat}(\textit{\textbf{R}}, \mathcal{A}(\textit{\textbf{R}}, \textit{\textbf{F}}_c))),
\end{equation}
where $\mathcal{A}(\cdot,\cdot)$ denotes cross-attention, and $\textit{\textbf{R}}_{Enh}$ has the same dimensions as the original features. 
Through interactions between original features and the category-aware features, the enhanced features contain higher-quality feature information. This makes the objects easier to classify and locate, enabling the detector to recognize objects more accurately.


\subsection{\qiankun{Integrating MCSS and CFEM into MAFE R-CNN}}
\label{MAFE}

As described in Section~\ref{MCSS} and Section~\ref{CFEM}, the MCSS and CFEM are designed to enhance the ability of detector to detect small objects by selecting more positive samples for small objects and enhancing small object features through learning category-aware features among objects. We elegantly integrate MCSS and CFEM into MAFE R-CNN, as shown in Figure~\ref{model_structure}(c).
First, we use a standard RoI head to perform the initial stage of regression and classification on the region of interest features, yielding initial predictions. Then, MCSS adaptively selects positive samples for subsequent stages of regression. Before the second-stage prediction, CFEM is used to enhance the candidate box features, and the enhanced features $R_{Enh}$ are used for regression and classification. 
Finally, the standard RoI head refines the predictions from the previous stage to produce the final detection results.
It is important to note that the output of each stage undergoes feature extraction and alignment for usage in the subsequent stage. MCSS and category-aware memory updates are only executed during the training phase, as ground truth information is not available during inference.

Unlike most small object detection methods, the proposed method does not require complex loss functions for constraint. Instead, it only utilizes the classification and regression losses commonly employed in traditional object detectors to ensure model robustness. Therefore, only includes the traditional classification and regression losses commonly used in our method:
\begin{equation}
    \mathcal{L} = \mathcal{L}_{cls} + \mathcal{L}_{reg},
\end{equation}
where $\mathcal{L}_{cls}$ and $\mathcal{L}_{reg}$ represent the cross-entropy loss and the smooth L1 loss~\cite{girshick2015fast}, respectively.

\begin{table*}[t]
\setlength{\tabcolsep}{7pt}
\renewcommand{\arraystretch}{1.3}
\caption{Comparison with different methods on the SODA-D benchmark. Two-stage, single-stage, anchor-free, and transformer-based methods are all evaluated. Except for YOLOX (CSPDarknet)~\cite{ge2021yolox} and CornerNet (HourglassNet-104)\cite{law2018cornernet}, all methods use ResNet-50~\cite{he2016deep} as the backbone. The best and second-best results are highlighted in bold and underlined, respectively.}
\label{table_main_results_sodad}
\centering
\begin{tabular}{l|c|c|c|ccc|cccc|c}
\hline
\hline
\multirow{1}{*}{\makecell{Method}} & \multirow{1}{*}{\makecell{Paradigm}} & \multirow{1}{*}{Publication} & \multirow{1}{*}{Epoch} & $AP$ & $AP_{50}$ & $AP_{75}$ & \multicolumn{1}{c}{$AP_{eS}$} & \multicolumn{1}{c}{$AP_{rS}$} & \multicolumn{1}{c}{$AP_{gS}$} & \multicolumn{1}{c|}{$AP_{N}$} & AR\\
\hline
CornerNet~\cite{law2018cornernet} & \multirow{3}{*}{\makecell{anchor free}} & ECCV’18 & 24 & 24.6 & 49.5 & 21.7 & 6.5 & 20.5 & 32.2 & 43.8 & 29.7\\
CenterNet~\cite{duan2019centernet} && ArXiv’19 & 70 & 21.5 & 48.8 & 15.6 & 5.1 & 16.2 & 29.6 & 42.4 & 23.4\\
RepPoints~\cite{yang2019reppoints} && ICCV’19 & 12 & 28.0 & 55.6 & 24.7 & 10.1 & 23.8 & 35.1 & 45.3 & 40.3\\
\hline
Deformable-DETR~\cite{zhu2020deformable} & \multirow{3}{*}{\makecell{Transformer}} & ICLR’20 & 50 & 19.2 & 44.8 & 13.7 & 6.3 & 15.4 & 24.9 & 34.2 & 21.7\\
Sparse RCNN~\cite{sun2021sparse} && CVPR’21 & 12 & 24.2 & 50.3 & 20.3 & 8.8 & 20.4 & 30.2 & 39.4 & 32.1\\
RT-DETR~\cite{zhao2024detrs} && {CVPR’24} & 72 & 26.7 & 50.7 & 24.1 & 10.2 & 22.5 & 30.9 & 39.6 & 31.0\\
\hline
RetinaNet\cite{lin2017focal} & \multirow{6}{*}{\makecell{single-stage}} & ICCV’17 & 12 & 28.2 & 57.6 & 23.7 & 11.9 & 25.2 & 34.1 & 44.2 & 40.6\\
FCOS~\cite{tian2019fcos} & & ICCV’19 & 12 & 23.9 & 49.5 & 19.9 & 6.9 & 19.4 & 30.9 & 40.9 & 27.7\\ 
ATSS~\cite{zhang2020bridging} & & CVPR'20 & 12 & 26.8 & 55.6 & 22.1 & 11.7 & 23.9 & 32.2 & 41.3 & 33.1\\
YOLOX~\cite{ge2021yolox} & & ArXiv’21 & 70 & 26.7 & 53.4 & 23.0 & 13.6 & 25.1 & 30.9 & 30.4 & 33.5\\
DyHead~\cite{dai2021dynamic} & & CVPR’21 & 12 & 27.5 & 56.1 & 23.2 & 12.4 & 24.4 & 33.0 & 41.9 & 38.7\\
\hline
Faster R-CNN~\cite{ren2015faster} & \multirow{6}{*}{\makecell{two-stage}} & NeurIPS’15 & 12 & 28.9 & 59.4 & 24.1 & 13.8 & 25.7 & 34.5 & 43.0 & 41.2\\
Cascade RPN~\cite{vu2019cascade} && NeurIPS’19 & 12 & 29.1 & 56.5 & 25.9 & 12.5 & 25.5 & 35.4 & 44.7 & 41.8\\
RFLA~\cite{xu2022rfla} && ECCV’22 & 12 & 29.7 & 60.2 & 25.2 & 13.2 & 26.9 & 35.4 & 44.6 & 40.8\\
Cascade R-CNN~\cite{cai2019cascade} && TPAMI’21 & 12 & \underline{31.2} & 59.4 & \underline{27.9} & 12.8 & {27.6} & \underline{37.8} & \underline{47.8} &\underline{43.7}\\
CFINet~\cite{yuan2023small} && ICCV'23 & 12 & 30.7 & \underline{60.8} & 26.7 & \underline{14.7} & \underline{27.8} & 36.4 & {44.6} & 42.6\\
KLDet~\cite{zhou2024kldet} && TGRS'24 & 12 & 25.9 & 53.8 & 21.4 & 10.7 & 22.2 & 31.9 & 41.6 & 35.9\\
\hline
MAFE R-CNN (Ours) & \multirow{1}{*}{\makecell{two-stage}}& -& 12& \bf{32.7}& \bf{61.4}& \bf{29.6}& \bf{15.3}& \bf{29.3}& \bf{38.9}& \bf{48.1}& \bf{46.1}\\
\hline
\hline
\end{tabular}
\end{table*}

\begin{table*}[!htbp]
\setlength{\tabcolsep}{7pt}
\renewcommand{\arraystretch}{1.3}
\caption{Comparison with different methods on the SODA-A benchmark. Two-stage and single-stage methods are all evaluated. The best and second-best results are highlighted in bold and underlined, respectively.}
\label{table_main_results_sodaa}
\centering
\begin{tabular}{l|c|c|c|ccc|cccc|c}
\hline
\hline
\multirow{1}{*}{\makecell{Method}} & \multirow{1}{*}{\makecell{Paradigm}} & \multirow{1}{*}{Publication} & \multirow{1}{*}{Epoch} & $AP$ & $AP_{50}$ & $AP_{75}$ & \multicolumn{1}{c}{$AP_{eS}$} & \multicolumn{1}{c}{$AP_{rS}$} & \multicolumn{1}{c}{$AP_{gS}$} & \multicolumn{1}{c|}{$AP_{N}$} & AR\\
\hline
Rotated RetinaNet\cite{lin2017focal} & \multirow{4}{*}{\makecell{single-stage}} & ICCV’17 & 12 & 26.8 & 63.4 & 16.2 & 9.1 & 22.0 & 35.4 & 28.2 & 40.7\\
$\text{S}^2$A-Net~\cite{han2021align} & & TGRSV’21 & 12 & 28.3 & 69.6 & 13.1 & 10.2 & 22.8 & 35.8 & 29.5 & 41.3\\ 
Oriented RepPoints~\cite{li2022oriented} & & CVPR'22 & 12 & 26.3 & 58.8 & 19.0 & 9.4 & 22.6 & 32.4 & 28.5 & -\\
DHRec~\cite{nie2022multi} & & TPAMI’22 & 12 & 30.1 & 68.8 & 19.8 & 10.6 & 24.6 & 40.3 & 34.6 & -\\
\hline
Faster R-CNN~\cite{ren2015faster} & \multirow{6}{*}{\makecell{two-stage}} & NeurIPS’15 & 12 & 32.5 & 70.1 & 24.3 & 11.9 & 27.3 & 42.2 & 34.4 & 42.8\\
Gliding Vertex~\cite{xu2020gliding} && TPAMI’20 & 12 & 31.7 & 70.8 & 24.3 & 11.9 & 27.3 & 42.2 & 34.4 & 42.4\\
Oriented R-CNN~\cite{xie2024oriented} && IJCV’24 & 12 & {34.4} & 70.7 & {28.6} & 12.5 & {28.6} & \underline{44.5} & \underline{36.7} &44.6\\
DODet~\cite{9706434} && TGRS’22 & 12 & 31.6 & 68.1 & 23.4 & 11.3 & 26.3 & 41.0 & 33.5 & -\\
CFINet~\cite{yuan2023small} && ICCV'23 & 12 & {34.4} & \underline{73.1} & 26.1 & {13.5} & {29.3} & 44.0 & {35.9} & -\\
LSKNet~\cite{li2024lsknet} && arXiv'24 & 12& \underline{35.5}& \bf{73.5}& \underline{29.5}& \underline{13.7}& \bf{32.3}& 41.5& \bf{37.5}& 44.5\\
\hline
MAFE R-CNN (Ours) & \multirow{1}{*}{\makecell{two-stage}}& -& 12& \bf{35.8}& \bf{73.5}& \bf{29.6}& \bf{13.9}& \underline{30.5}& \bf{45.8}& \bf{37.5}& \bf{44.7}\\
\hline
\hline
\end{tabular}
\end{table*}

\section{Experiments}

In this section, we first introduce the dataset and implement details. Then, we compare the proposed method with other detectors and present ablation analyses.

\subsection{Dataset and Settings}


We evaluate our method on the large-scale benchmark dataset SODA~\cite{cheng2023towards}. The dataset is divided into two subsets based on autonomous driving and remote sensing scenarios, \textit{i.e.}, SODA-D and SODA-A.

\noindent \textbf{SODA-D} primarily focuses on small objects in driving scenarios, encompassing nine categories: \textit{people, rider, bicycle, motor, vehicle, traffic-sign, traffic-light, traffic-camera}, and \textit{warning cones.} 
The dataset has 24,828 images captured in various lighting and weather conditions, with 278,433 instances.

\noindent \textbf{SODA-A} focuses on remote sensing scene, containing 2,513 images and 872,069 instances with oriented bounding box annotations. 
Instances can appear at any angle, with variations in both dense and sparse regions. The dataset covers nine categories: \textit{airplane, helicopter, small vehicle, large vehicle, ship, container, storage tank, swimming pool}, and \textit{windmill.}

Compared to traditional object detection datasets where most objects are larger than 1024 pixels, as a large-scale benchmark for small object detection, SODA contains a large number of small objects with average sizes ranging from 10 to 400 pixels. This makes it a considerably challenging benchmark for SOD.

\noindent \textbf{Metrics.}
The metric for SODA adheres to the commonly used evaluation standard. 
In terms of object region categorization, we follow the evaluation criteria proposed in \cite{yuan2023small} to compute the Average Precision (AP) and Average Recall (AR) for small object detection. We categorize small instances into three groups based on their region areas: extremely Small (eS), relatively Small (rS), and generally Small (gS). The specific area delineations are as follows: (0, 144], (144, 400], and (400, 1024]. Other objects with areas exceeding 1024 pixels are classified as Normal objects.

\noindent \textbf{Implement Details.} 
Considering the high resolution of images in the SODA dataset (4000 $\times$ 3000), we follow the method in \cite{xia2018dota} to divide the original images into multiple patches of size 800 $\times$ 800, with a stride of 650. During training and inference, images are resized to 1200 $\times$ 1200 when fed into the network.
Following two-stage detectors~\cite{ren2015faster}, we adopt ResNet50~\cite{he2016deep} as the backbone, along with FPN~\cite{lin2017feature}.
In MCSS, we set $k$ to 9, category confidence weight $\alpha$ to 0.3, and minimum positive sample threshold $\beta$ to 0.6. Following~\cite{cheng2023towards,lin2014microsoft}, we set $S_{max}$ to 32. Additionally, based on experimental observations, for objects with an absolute size of 32, the standard scale $\gamma$ is set to 1.0, and the maximum $\gamma$ is 3.0. 
The dimension $D$ of the category-aware memory is set to 1024. Category-aware memory $\textit{\textbf{M}}$ are initialized by a normal distribution with the same random seed.
We train our model 12 epochs for SODA-A and SODA-D on two RTX 3090 GPUs. The batch sizes are set to 8 and 4, and the learning rates to 0.04 and 0.01, respectively.
The learning rate decays after epoch 8 and epoch 11 by 0.1.
The default optimizer is SGD with the momentum of 0.9 and the weight decay of 0.0001.

\noindent \textbf{Comparison Method.} 
We compare MAFE R-CNN with state-of-the-art object detectors, including four detection architectures: anchor-free detectors (CornerNet~\cite{law2018cornernet}, CenterNet~\cite{duan2019centernet}, and RepPoints~\cite{yang2019reppoints}), transformer-based detectors (Deformable-DETR~\cite{zhu2020deformable}, Sparse RCNN~\cite{sun2021sparse}, and RT-DETR~\cite{zhao2024detrs}), single-stage detectors (RetinaNet~\cite{lin2017focal}, FCOS~\cite{tian2019fcos}, ATSS~\cite{zhang2020bridging}, YOLOX~\cite{ge2021yolox}, and DyHead~\cite{dai2021dynamic}), and two-stage detectors (Faster R-CNN~\cite{ren2015faster}, Cascade RPN~\cite{vu2019cascade}, RFLA~\cite{xu2022rfla}, Cascade R-CNN~\cite{cai2019cascade}, CFINet~\cite{yuan2023small}, and KLDet~\cite{zhou2024kldet}).


\begin{figure*}[t]
	\centering
	\includegraphics[width=\linewidth, height=10cm]{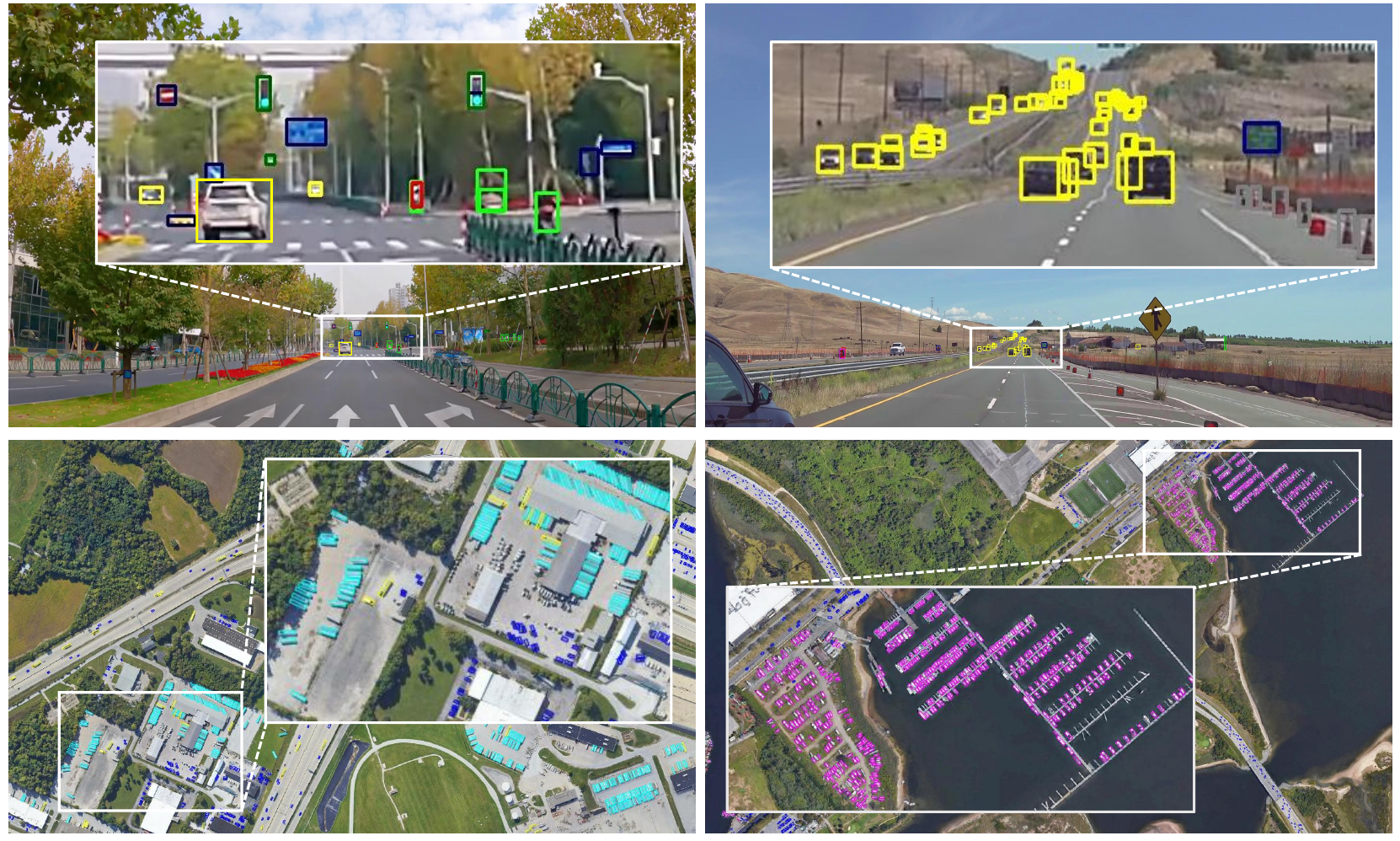}
	\caption{Visualization on the SODA test set. Only small objects with a confidence score larger than 0.3 are shown. Best viewed in color and zoom-in windows.}
	\label{visualization}
\end{figure*}

\subsection{Main Results}

To demonstrate the effectiveness of the proposed MAFE R-CNN, we compare it with previous methods on the SODA-D and SODA-A. Table~\ref{table_main_results_sodad} and Table~\ref{table_main_results_sodaa} show the experimental results of the proposed method in comparison with several commonly used detectors on the SODA dataset. Given the specific requirement in remote sensing to predict rotated bounding boxes, we compare the proposed method with commonly used remote sensing detection frameworks.

\subsubsection{Comparison with previous state-of-the-art methods}
As shown in Table~\ref{table_main_results_sodad}, MAFE R-CNN achieves $32.7\%$ AP and $46.1\%$ AR on SODA-D, surpassing all previous small object detection methods, including ATSS~\cite{zhang2020bridging}, KLDet~\cite{zhou2024kldet}, RFLA~\cite{xu2022rfla}, and CFINet~\cite{yuan2023small}. 
\qiankun{It is} worth noting that MAFE R-CNN performs better than the second-ranked Cascade-RCNN, with an AP increase of $1.5\%$ and an AR increase of $2.4\%$, and an average AP improvement of more than $1.1\%$ in small object detection.
As shown in Table~\ref{table_main_results_sodaa}, MAFE R-CNN achieves $35.8\%$ AP and $44.7\%$ AR on SODA-A, surpassing previous remote sensing methods, including LSKNet~\cite{li2024lsknet}, CFINet~\cite{yuan2023small}, Oriented R-CNN~\cite{xie2024oriented}, and Faster R-CNN~\cite{ren2015faster}. 
Figure~\ref{visualization} shows the visual detection results of our method on the SODA test dataset, intuitively showing the capability of the proposed method in the small object detection task.

\subsubsection{Comparison with various detectors}
In Tables~\ref{table_main_results_sodad} and~\ref{table_main_results_sodaa}, we categorize detectors into anchor-free detectors, transformer-based detectors, single-stage detectors, and two-stage detectors.
For anchor-free detectors like CornerNet~\cite{law2018cornernet} and CenterNet~\cite{duan2019centernet}, the lack of anchor constraints makes it easy to lose information about small objects during feature extraction. Consequently, they fail to generate corresponding anchor points, resulting in poor detection performance for small objects.
Regarding transformer detectors, although they consider global information in the image through self-attention mechanisms, they are not sensitive enough to small object information, greatly limiting small object detection performance, as elaborated in~\cite{carion2020end}.
As for single-stage detectors~\cite{ge2021yolox,tian2019fcos,lin2017focal,zhang2020bridging,dai2021dynamic}, which perform only one-stage prediction, they lack an iterative process compared to two-stage detectors. In small object detection tasks, the limited number of candidate boxes and less prominent features exacerbate this limitation, making it challenging for single-stage detectors to obtain precise predictions. 
Compared to other detection methods, two-stage detectors achieve relatively accurate results in small object detection through multiple prediction stages. However, imbalanced sample selection and blurred small object features still limit the performance of detectors in small object detection.


\subsection{Discussions}
\label{ablation_study}

We first discuss the individual contributions of the proposed core components and the combined network. Then, we further analyze the experimental results by examining the hyperparameters in MCSS and the key steps in CFEM. Unless otherwise specified, all experiments below are conducted on the SODA-D dataset.

\subsubsection{Discussions on MAFE R-CNN} 
Table~\ref{table_ablation_main} shows a comprehensive comparison experiment, meticulously assessing the individual contributions of Multi-Clue Sample Selection (MCSS) and Category-aware Feature Enhancement Module (CFEM) in the final prediction process of the Multi-clue Assignment and Feature Enhancement (MAFE R-CNN). 
Within MAFE R-CNN, MCSS is utilized to select high-quality positive samples, while CFEM enhances object features. 
Row 1 in Table~\ref{table_ablation_main} represents the original Cascade RoI heads from Cascade R-CNN~\cite{cai2019cascade}. As observed, both MCSS and CFEM demonstrate improved performance in small object detection. Specifically, MCSS and CFEM respectively increase the $AP$ by 1.0\% and 0.8\%, with enhancements of 1.0\% or more in small object $AP_{eS}$ and $AP_{gS}$. 
Furthermore, the combination of MCSS and CFEM demonstrates a synergistic effect, resulting in substantial performance gains, which highlight the complementary nature of sample selection and feature enhancement, achieving the best detection results. 

As shown in Table~\ref{table_ablation_cadh_stage}, we further investigated the impact of applying MAFE R-CNN to different stages of the RoI head, specifically by integrating MCSS and CFEM into various RoI heads during training and testing.
The results indicate that MAFE R-CNN in either the first or third stage leads to a decrease in detection performance. Specifically, classification and regression of the first stage help produce higher-quality candidate samples, which better support learning and prediction in the second stage. Conversely, applying MAFE R-CNN in the third stage is less effective, as small object samples have often been filtered out in the first two stages.

\begin{table}[t]
\setlength{\tabcolsep}{2.5pt}
\renewcommand{\arraystretch}{1.3}
\caption{Results on core components of MAFE R-CNN}
\label{table_ablation_main}
\centering
\begin{tabular}{cc|ccc|cccc|c}
\hline
\hline
\multirow{1}*{MCSS} &\multirow{1}*{CFEM} & \multirow{1}*{$AP$} & \multirow{1}*{$AP_{50}$} & \multirow{1}*{$AP_{75}$} & \multirow{1}*{$AP_{eS}$}& \multirow{1}*{$AP_{gS}$}& \multirow{1}*{$AP_{rS}$}& \multirow{1}*{$AP_{N}$}& \multirow{1}*{$AR$} \\
\hline
& & {31.2} & 59.4 & {27.9} & 12.8 & {27.6} & {37.8} & {47.8} &43.7\\
\checkmark &  &{32.2}& {60.6}& {29.0}& {14.8}& {28.8}& {38.5}& {48.4}& {43.9}\\
 & \checkmark& {32.0}& {60.2}& {29.1} &{15.0}& {28.7}& {38.2}& \bf{48.5}& {43.7}\\
\checkmark& \checkmark& \bf{32.7}& \bf{61.4}& \bf{29.6}& \bf{15.3}& \bf{29.3}& \bf{38.9}& {48.1}& \bf{46.1} \\

\hline
\hline
\end{tabular}
\end{table}

\begin{table}[t]
\setlength{\tabcolsep}{3pt}
\renewcommand{\arraystretch}{1.3}
\caption{Results with MAFE R-CNN in different stage of RoI head, where $S_x$ denotes the $x^{th}$ stage.}
\label{table_ablation_cadh_stage}
\centering
\begin{tabular}{ccc|ccc|cccc|c}
\hline
\hline
    \multicolumn{3}{c|}{MAFE R-CNN} & \multirow{2}*{$AP$} & \multirow{2}*{$AP_{50}$} & \multirow{2}*{$AP_{75}$} & \multirow{2}*{$AP_{eS}$} & \multirow{2}*{$AP_{rS}$} & \multirow{2}*{$AP_{gS}$} & \multirow{2}*{$AP_{N}$} & \multirow{2}*{$AR$} \\
\cline{1-3}
$S_1$& $S_2$& $S_3$& & & & & & & \\
\hline
 \checkmark & & & 32.3& 60.8& 29.3& 15.1& 29.0& 38.5& 47.8& 45.6 \\
 & \checkmark& &\bf{32.7}& \bf{61.4}& \bf{29.6}& \bf{15.3}& \bf{29.3}& \bf{38.9}& \bf{48.1}& \bf{46.1}\\
 & & \checkmark & 32.1& 60.7& 29.1& 15.1& 28.7& 38.2& 47.6& 45.5\\
\hline
\hline
\end{tabular}
\end{table}

\begin{table}[!thbp]
\setlength{\tabcolsep}{1pt}
\renewcommand{\arraystretch}{1.3}
\caption{The results of different numbers of RoI heads. Except for stage number equal to 1, and MAFE R-CNN is directly used, all other experiments default to using MAFE R-CNN in the second stage}
\label{table_ablation_head_stages}
\centering
\begin{tabular}{c|c|ccc|cccc|c}
\hline
\hline
\multirow{2}*{\makecell[c]{Heads \\ Num}} &\multirow{2}*{Method} &\multirow{2}*{$AP$} & \multirow{2}*{$AP_{50}$} & \multirow{2}*{$AP_{75}$} & \multirow{2}*{$AP_{eS}$}& \multirow{2}*{$AP_{gS}$}& \multirow{2}*{$AP_{rS}$}& \multirow{2}*{$AP_{N}$}& \multirow{2}*{$AR$} \\
& & & & & & & & &\\
\hline
1& Faster RCNN\cite{ren2015faster}& 28.9 & 59.4 & 24.1 & 13.8 & 25.7 & 34.5 & 43.0 & 41.2\\
1& CFINet\cite{yuan2023small}& 30.4 & {60.1} & 26.6 &  {14.7} & 27.4 & 36.0 & {44.9} & 41.9\\
\hline
1& \scriptsize{MAFE R-CNN(Ours)}& 31.0& 60.7& 27.0& 14.5& 28.0& 36.8& 45.3& 42.9\\
2& \scriptsize{MAFE R-CNN(Ours)}& 31.5& 61.2& 27.9& 14.5& 28.4& 37.3& 45.8& 43.7\\
3& \scriptsize{MAFE R-CNN(Ours)}& \bf{32.7}& \bf{61.4}& \bf{29.6}& \bf{15.3}& \bf{29.3}& \bf{38.9}& \bf{48.1}& \bf{46.1}\\
4& \scriptsize{MAFE R-CNN(Ours)}& 32.4& 60.2& 29.8& 14.9& 28.7& 38.9& 48.3& 45.1\\
\hline
\hline
\end{tabular}
\end{table}

To demonstrate the effectiveness of multi-stage RoI heads, we analyze the impact of different numbers of RoI heads on detection results in Table~\ref{table_ablation_head_stages}.
Notably, we integrate MAFE R-CNN into the second stage RoI head for these experiments.
Rows 1-3 of Table~\ref{table_ablation_head_stages} show the experimental results using a single RoI head. It can be observed that directly using MAFE R-CNN achieves the best results compared to other single RoI head methods. 
Rows 4-6 show the results for multi-stage RoI heads. 
Clearly, the cascading of RoI heads results in better detection performance, as shown in the $5^{th}$ row, the integration of three-stage RoI heads performs the best results, 
with an increase of {1.7\%} in $\text{AP}$ and {3.2\%} in $\text{AR}$ compared to single RoI head. 
Furthermore, the results in the $5^{th}$ and $6^{th}$ rows indicate that the cascading of three RoI heads already achieves satisfactory performance.
More RoI heads do not achieve better detection results but will affect the inference speed of the detector.
It is worth noting that the multi-stage RoI heads do not affect the model's parameters or FLOPs significantly. For example, increasing from one RoI head to three RoI heads results in only a 5\% increase in parameters.

\subsubsection{Discussions on MCSS.} 
MCSS effectively improves the performance of small object detection, as shown in the $2^{nd}$ row of Table~\ref{table_ablation_main}. 
We conducted detailed experiments to analyze the maximum number of positive samples and the category confidence weight in MCSS, \textit{i.e.}, $k$ and $\alpha$.
It is important to note that, to ensure fairness, CFEM is not used in the experiments shown in Table~\ref{table_ablation_k} and Table~\ref{table_ablation_a}, and MCSS is applied by default to the second RoI head.

As shown in Table~\ref{table_ablation_k}, we conduct an analysis experiment using the parameter set $k \in \{3,6,9,12,15,18\}$.
MCSS achieves best performance when $k$ is set to 9. 
Specifically, when $k$ is too small (\textit{e.g.}, $k$=3), MCSS fails to select a sufficient number of samples, leading to a decrease in $\text{AP}$ by 0.4\% compared to $k$=9. 
On the other hand, when $k$ is too large (\textit{e.g.}, $k$=18), MCSS introduces a large number of low-quality samples, resulting in an $\text{AP}$ drop of 1.0\% compared to $k$=9. 
Therefore, both the quantity and quality of samples impact the detector's training results.

Similarly, we train MCSS using different category confidence weights $\alpha$. 
A larger value of $\alpha$ indicates a higher importance of category confidence in sample selection.
As shown in Table~\ref{table_ablation_a}, MCSS achieves the best performance when $\alpha$ is set to 0.3. 
Compared to not considering category confidence at all (the $1^{st}$ row in Table~\ref{table_ablation_a}), using multi-clue confidence sample selection yields better results, with an increase of 0.5\% in $\text{AP}$, and more importantly, an improvement of 1.0\% in $\text{AP}_{eS}$. 
Additionally, when $\alpha$ increases, even when $\alpha$=1, MCSS does not exhibit a significant decrease in performance, as we constrain candidate samples using center distance and parameter $k$ before sample selection, enhancing the stability.


\begin{table}[t]
\setlength{\tabcolsep}{4pt}
\renewcommand{\arraystretch}{1.3}
\caption{Results with different value of $k$.}
\label{table_ablation_k}
\centering
\begin{tabular}{c|ccc|cccc|c}
\hline
\hline
\multirow{1}{*}{$k$} & $AP$ & $AP_{50}$ & $AP_{75}$ & \multicolumn{1}{c}{$AP_{eS}$} & \multicolumn{1}{c}{$AP_{rS}$} & \multicolumn{1}{c}{$AP_{gS}$} & \multicolumn{1}{c|}{$AP_{N}$} & \multicolumn{1}{c}{$AR$}\\
\hline
3& 31.8& 59.9& 28.9& 14.6& 28.4& 38.1& 48.2& 43.5\\
6& 32.0& 60.4& 28.9& 14.6& 28.6& 38.5& 48.3& 43.8\\
9& \bf{32.2}& {60.6}& \bf{29.0}& \bf{14.8}& \bf{28.8}& \bf{38.5}& \bf{48.4}& {43.9}\\
12& 32.1& \bf{60.8}& 28.9& 14.7& \bf{28.8}& 38.4& 48.2& \bf{44.0}\\
15& 31.8& 60.5& 28.7& 14.4& 28.6& 38.0& 47.9& 43.7\\
18& 31.2& 59.9& 27.9& 14.2& 27.8& 37.3& 46.7& 42.5\\
\hline
\hline
\end{tabular}
\end{table}

\begin{table}[t]
\setlength{\tabcolsep}{4pt}
\renewcommand{\arraystretch}{1.3}
\caption{Results with different category confidence weight $\alpha$.}
\label{table_ablation_a}
\centering
\begin{tabular}{c|ccc|cccc|c}
\hline
\hline
\multirow{1}{*}{$\alpha$} & $AP$ & $AP_{50}$ & $AP_{75}$ & \multicolumn{1}{c}{$AP_{eS}$} & \multicolumn{1}{c}{$AP_{rS}$} & \multicolumn{1}{c}{$AP_{gS}$} & \multicolumn{1}{c|}{$AP_{N}$} & \multicolumn{1}{c}{$AR$}\\
\hline
0& 31.7& 60.0& 28.1& 13.8& 27.3& 38.2& \bf{48.8}& 43.4\\
0.3& \bf{32.2}& \bf{60.6}& \bf{29.0}& \bf{14.8}& \bf{28.8}& \bf{38.5}& {48.4}& \bf{43.9}\\
0.5& {32.0}& {60.2}& {28.9}& 14.7& {28.7}& {38.2}& 48.5& 43.7\\
0.7& 32.1& 60.4& 29.0& 14.6& 28.7& 38.4& 48.1& 43.8\\
0.9& 32.1& \bf{60.6}& 29.0& 14.6& 28.5& 38.4& 48.1& 43.5\\
1.0& 32.1& 60.4& 29.2& 14.7& 28.5& 38.5& 48.5& 43.7\\
\hline
\hline
\end{tabular}
\end{table}

\subsubsection{\qiankun{Discussions on CFEM}}
In Table~\ref{table_ablation_main}, we demonstrate the effectiveness of CFEM for small object detection. We then conduct comprehensive studies on its key steps. To ensure fairness, we do not use MCSS in these experiments.

As shown in Table~\ref{table_ablation_cfem}, we investigate different feature information and the attention mechanism in CFEM individually.
The enhanced features contain more category-aware information, which is beneficial for detection. However, using only $\textit{\textbf{F}}_{Enh}$ for prediction without concatenating the candidate box features $\textit{\textbf{R}}$ results in the loss of feature information, and may even lead to worse performance. 
The comparison between the experimental results in rows $3^{rd}$ and $4^{th}$ demonstrates that cross-attention plays a crucial role in mining the potential relationships between candidate box features and category-aware memory features.
Furthermore, we further analyze the impact of CFEM on regression and classification in Table~\ref{table_ablation_cfem_1}. 
Due to the high-quality features provided by CFEM, applying CFEM to either classification or regression can improve detection performance to some extent. 
Therefore, we utilize the enhanced features in both classification and regression to achieve the best results. This indicates that the proposed CFEM effectively learns high-quality category-aware features to assist in classification and regression in object detection.

\begin{table}[t]
\setlength{\tabcolsep}{2pt}
\renewcommand{\arraystretch}{1.3}
\caption{Results with key steps of CFEM. The best results is highlighted in bold.}
\label{table_ablation_cfem}
\centering
\begin{tabular}{ccc|ccc|cccc|c}
\hline
\hline
\multirow{1}*{$\textit{\textbf{R}}$} &\multirow{1}*{$\textit{\textbf{F}}_{\scriptsize{Enh}}$} & \multirow{1}*{$\mathcal{A}(\cdot,\cdot)$} & \multirow{1}*{$AP$} & \multirow{1}*{$AP_{50}$} & \multirow{1}*{$AP_{75}$} & \multirow{1}*{$AP_{eS}$}& \multirow{1}*{$AP_{gS}$}& \multirow{1}*{$AP_{rS}$}& \multirow{1}*{$AP_{N}$}& \multirow{1}*{$AR$} \\
\hline
\checkmark & & & {31.2} & 59.4 & {27.9} & 12.8 & {27.6} & {37.8} & {47.8} & 43.7\\
& \checkmark& &   30.7& 59.0& 27.6& 13.9& 27.3& 37.3& 46.8& 43.5\\
\checkmark & \checkmark& & 31.8& 60.1& 28.3& 14.4& 28.1& \bf{38.2}& 48.2& 43.5 \\
\checkmark & \checkmark& \checkmark& \bf{32.0}& \bf{60.2}& \bf{29.1} &\bf{15.0}& \bf{28.7}& \bf{38.2}& \bf{48.5}& \bf{43.7}  \\

\hline
\hline
\end{tabular}
\end{table}

\begin{table}[!htbp]
\setlength{\tabcolsep}{3pt}
\renewcommand{\arraystretch}{1.2}
\caption{Results with CFEM for classification and regression.}
\label{table_ablation_cfem_1}
\centering
\begin{tabular}{cc|ccc|cccc|c}
\hline
\hline
\multirow{1}*{Reg.} &\multirow{1}*{Cls.} & \multirow{1}*{$AP$} & \multirow{1}*{$AP_{50}$} & \multirow{1}*{$AP_{75}$} & \multirow{1}*{$AP_{eS}$}& \multirow{1}*{$AP_{gS}$}& \multirow{1}*{$AP_{rS}$}& \multirow{1}*{$AP_{N}$}& \multirow{1}*{$AR$} \\
\hline
\checkmark & & 31.8& 59.3& 28.4& 14.6& 28.5& \bf{38.2}& 48.1& 43.5 \\
 & \checkmark& 31.8& 59.7& 28.1& 14.4& 28.3& 38.0& \bf{48.4}& 43.2 \\
\checkmark& \checkmark& \bf{32.0}& \bf{60.2}& \bf{29.1} &\bf{15.0}& \bf{28.7}& \bf{38.2}& {48.1}& \bf{43.7}\\

\hline
\hline
\end{tabular}
\end{table}

\begin{figure}[!htbp]
    \centering
    \includegraphics[width=\linewidth]{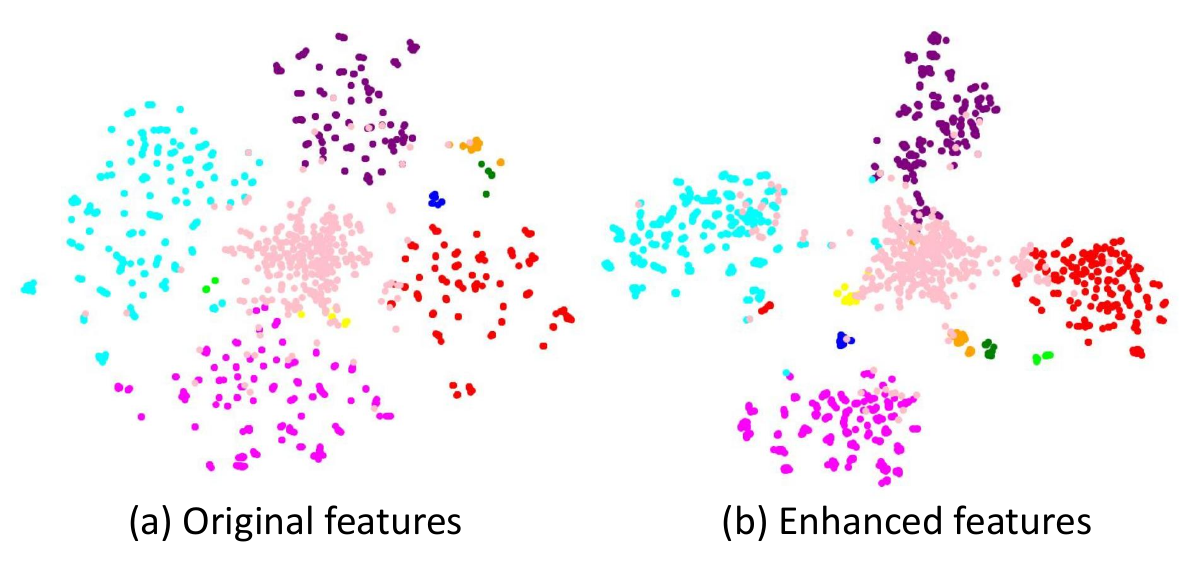}
    \caption{Comparison of proposal feature distributions before (a) and after (b) feature enhancement. Different colors represent different categories.}
    \label{proposal_distribution}
\end{figure}

Finally, we visualize the feature distributions before and after feature enhancement.
Figure~\ref{proposal_distribution} illustrates the comparison of candidate box feature distributions before and after interaction with category-aware features. 
Different colors represent the features of different categories. It can be observed that after aggregating category-aware feature information, the features of candidate boxes within the same category exhibit a more consolidated distribution, demonstrating the effectiveness of feature enhancement.

\section{Conclusion}

    In this paper, we focus on the task of small object detection and propose MAFE R-CNN that enhances both the quantity and quality of small object samples. This method ingeniously integrates multi-clue sample selection and a category-aware feature enhancement mechanism into the detector, utilizing a multi-stage refinement approach for classification and regression. The former provides a more balanced and higher-quality set of positive samples for model training, while the latter enhances the representation of small object features through category-aware information. The proposed method effectively addresses the issues of imbalanced samples and blurred features in small object detection. Experimental results demonstrate that our method achieves state-of-the-art performance on the large-scale small object detection benchmark SODA.
    In future work, it is worth further exploring small object detection methods that address the challenges of few-shot or imbalanced data, improving the model's robustness across various object categories to achieve more refined and generalized small object detection.




 
%
\normalem
\input{bare_jrnl_new_sample4.bbl}

\bibliographystyle{IEEEtran}
\bibliography{bare_jrnl_new_sample4}

\newpage

 




\vfill

\end{document}